\documentclass[10pt,twocolumn,letterpaper]{article}
\usepackage[belowskip=-15pt,aboveskip=0pt]{caption}
\usepackage{iccv}
\usepackage{times}
\makeatletter
\@namedef{ver@everyshi.sty}{}
\makeatother
\usepackage{tikz}
\usepackage{pgfplots}
\usepackage{epsfig}
\usepackage{graphicx}
\usepackage{amsmath}
\usepackage{amssymb}
\newcommand{\VV}{\mathcal{V}}

\usepackage{pifont}
\newcommand{\cmark}{\ding{51}}%
\newcommand{\xmark}{\ding{55}}%

\usepackage{float}

\usepackage{adjustbox}
\usepackage{tabularx}

\usepackage{pgfplots}
\pgfplotsset{compat = 1.11}
\usepgfplotslibrary{groupplots}

\usepackage{multirow}
\usepackage{xcolor}
\usepackage[linesnumbered,ruled,vlined]{algorithm2e}

\SetCommentSty{mycommfont}

\usepackage{graphicx}
\usepackage{subfig}

\usepackage[pagebackref=true,breaklinks=true,letterpaper=true,colorlinks,bookmarks=false]{hyperref}

\iccvfinalcopy 

\def\iccvPaperID{2647} 

\ificcvfinal\fi

\definecolor{amber}{rgb}{1.0, 0.75, 0.0}
\definecolor{amaranth}{rgb}{0.9, 0.17, 0.31}
\definecolor{amethyst}{rgb}{0.6, 0.4, 0.8}
\definecolor{airforceblue}{rgb}{0.36, 0.54, 0.66}
\definecolor{green(munsell)}{rgb}{0.0, 0.66, 0.47}
\definecolor{azure(colorwheel)}{rgb}{0.0, 0.5, 1.0}

\begin{document}

\title{ZS-IL: Looking Back on Learned Experiences \\ For Zero-Shot Incremental Learning }

\author{Mozhgan PourKeshavarz, Mohammad Sabokrou\\
Institute For Research In Fundamental Sciences(IPM)\\
}

\maketitle
\ificcvfinal\thispagestyle{empty}\fi

\begin{abstract}
   Classical deep neural networks are limited in their ability to learn from emerging streams of training data. When trained sequentially on new or evolving tasks, their performance degrades sharply, making them inappropriate in real-world use cases. Existing methods tackle it by either storing old data samples or only updating a parameter set of DNNs, which, however, demands a large memory budget or spoils the flexibility of models to learn the incremented class distribution. In this paper, we shed light on an on-call transfer set to provide past experiences whenever a new class arises in the data stream. In particular, we propose a Zero-Shot Incremental Learning not only to replay past experiences the model has learned but also to perform this in a zero-shot manner. Towards this end, we introduced a memory recovery paradigm in which we query the network to synthesize past exemplars whenever a new task (class) emerges. Thus, our method needs no fixed-sized memory, besides calls the proposed memory recovery paradigm to provide past exemplars, named a transfer set in order to mitigate catastrophically forgetting the former classes. Moreover, in contrast with recently proposed methods, the suggested paradigm does not desire a parallel architecture since it only relies on the learner network. Compared to the state-of-the-art data techniques without buffering past data samples, ZS-IL demonstrates significantly better performance on the well-known datasets (CIFAR-10, Tiny-ImageNet) in both Task-IL and Class-IL settings.
\end{abstract}
\vspace{-5mm}
\section{Introduction}
\vspace{-2mm}
Conventional Deep Neural Networks (DNNs) are typically trained in an offline batch setting where all data are available at once. However, in real-world settings, models may incrementally encounter multiple tasks when trained online. In such scenarios, deep learning models suffer from catastrophic forgetting\cite{mccloskey1989catastrophic}, meaning they forget the previously obtained knowledge when adapting to the new information from the incoming observations. This is mainly because, models overwrite the decisive parameters for earlier tasks while learning the new one. As a preliminary solution, one can buffer all the observations received from the beginning and start over when it is needed, but such method is computationally expensive, and impracticable for real-world applications. 

Incremental Learning (IL) methods aim at training a single DNN from an unlimited stream of data, mitigating catastrophic forgetting conditioned on the limited computational overhead and memory budget. In the IL literature, the typical setting is a model to learn many tasks sequentially where each task contains one or more disjoint classification problems. The majority of the existing works assume that task identities are provided at the test time so that one can select the relevant part of the network for each example. This setup has been named Task-IL, where a more general circumstance is that task labels are available only during training, has been named class-IL.
	\vspace{-2mm}
\begin{table*}[t]
\caption{Incremental Learning approaches and their major requirements to be executed.}
\scriptsize
\renewcommand{\arraystretch}{1.5}
\begin{tabular*}{1\textwidth}{c@{\extracolsep{\fill}}c@{\extracolsep{\fill}}c@{\extracolsep{\fill}}c@{\extracolsep{\fill}}c@{\extracolsep{\fill}}c@{\extracolsep{\fill}}c@{\extracolsep{\fill}}c@{\extracolsep{\fill}}c@{\extracolsep{\fill}}c@{\extracolsep{\fill}}c@{\extracolsep{\fill}}c@{\extracolsep{\fill}}c@{\extracolsep{\fill}}c@{\extracolsep{\fill}}c@{\extracolsep{\fill}}c@{\extracolsep{\fill}}c@{\extracolsep{\fill}}c@{\extracolsep{\fill}}c@{\extracolsep{\fill}}c@{\extracolsep{\fill}}c@{\extracolsep{\fill}}}
\multirow{2}{*}{\textbf{Methods}} & oEWC & ALASSO & LwF & PNN & UCB & DGM & DGR & MeRGAN & MER & a-GEM & iCaRL & FDR & GSS & HAL & DER & BIC & WA & Mnemo. & TPCIL & \textbf{ZS-IL}  \\ 
 & \cite{schwarz2018progress} & \cite{park2019continual} & \cite{li2017learning} & \cite{rusu2016progressive} & \cite{ebrahimi2019uncertainty} & \cite{ostapenko2019learning}  & \cite{shin2017continual} & \cite{wu2018memory} & \cite{riemer2018learning} & \cite{chaudhry2018efficient} & \cite{rebuffi2017icarl} & \cite{benjamin2018measuring} & \cite{aljundi2019gradient} & \cite{chaudhry2020using}  & \cite{buzzega2020dark} & \cite{wu2019large} &  \cite{zhao2020maintaining} &  \cite{liu2020mnemonics} & \cite{tao2020topology} &\textbf{Ours} \\ \hline 
\textbf{Auxiliary} \textbf{network} & \xmark & \xmark & \xmark & \xmark & \xmark & \cmark & \cmark & \cmark & \xmark & \xmark &  \xmark & \xmark & \xmark & \xmark & \xmark & \xmark & \xmark & \xmark & \cmark & \xmark\\ 
\hline
\textbf{Test} \textbf{time oracle} & \xmark & \xmark & \cmark & \cmark & \xmark & \cmark & \cmark & \cmark & \cmark & \cmark & \xmark & \xmark & \xmark & \cmark & \xmark & \xmark & \xmark & \xmark & \xmark & \xmark \\ \hline
\textbf{Memory} \textbf{buffer} & \xmark & \xmark & \xmark & \xmark&\xmark & \xmark & \xmark & \xmark & \xmark & \cmark & \cmark & \cmark & \cmark & \cmark &  \cmark& \cmark &\cmark  & \cmark &\cmark  & \xmark\\  \hline
\end{tabular*}
\label{tab:comparison}
\end{table*}

Recently, various approaches have been put forward to solve catastrophic forgetting. Some works investigate on parameter isolation approaches using either expanding network architecture\cite{aljundi2017expert,xu2018reinforced,yoon2017lifelong,rosenfeld2018incremental} or considering a fixed-sized set of model parameters\cite{masse2018alleviating}, pruning\cite{mallya2018packnet} to learn a new task. Thus, they are only applicable to the task incremental scenario. Besides, several studies suggest a regularization term added on the loss function\cite{kirkpatrick2017overcoming,aljundi2018memory,zenke2017continual,nguyen2017variational}  to preserve the knowledge of previous classes without a need for knowing the task label during inference. Despite the increasing number of these methodologies, they can not touch a reasonable performance. Another followed terminology is experience reply-based methods which  store actual data samples from the past\cite{buzzega2020dark,riemer2018learning,chaudhry2018efficient} to alleviate catastrophic forgetting problem, where more of the recent state-of-the-art(SOTA) works fall into this category. Despite the success of this approach, they suffer from a main limitation that is the fixed memory budget when increasing the number of classes, leading to severe performance degradation. To overcome this constraint, several works propose to generate past data samples\cite{zhai2019lifelong}, as pseudo-data examples, conditioned on the model underlying distribution. Although their method is considered a data-free approach by discarding the need of storing real data, they achieve better performance than previous data-free works. However, one main drawback that these approach encounter is a need for a high-capacity auxiliary network to generate data which increase the complexity of the system in terms of training-difficulty and computation overhead.

To address the mentioned problems, one promising solution could be the capability to replay past experiences in a zero-shot manner. In this paper, we propose a novel zero-shot incremental learning (ZS-IL) paradigm, which allows the model to remember previous knowledge by itself, thereby eliminating the need for external memory. Also, in the proposed method, there is no need for a parallel network to generate past exemplars. Besides, we ask the learner network to remember what it has learned before learning new classes. Inspired by \cite{fredrikson2015model,he2019model,nayak2019zero}, we propose  a  solution  to reply the previous experiences by recovery the learner network memory. The main idea is  having only the learned network to not only remember what it has learned but also stay learned while encountering newly received knowledge.  how humans continually learn and remember them by themselves with no external help. The most likely method to learn in this way is the regularization-based method, as compared in Table \ref{tab:comparison}. However, they still could not compromise between human-likely learning and performance. Besides, our method can achieve a better result by a significant margin with no additional requirement.

In summary, the main contributions of this paper are: (1)To the best of our knowledge, we introduce, for the first time, the idea of ZS-IL to make experience replay without a need to a parallel auxiliary network, or any other additional requirement, (2)Since we only rely on the learner network, there is no need for buffering past data samples, extracted prior knowledge, and a parallel high-capacity network, resulting in significantly reduce computation cost and memory footprint, and  (3)Our method compromises between decreasing memory budget and maintaining performance comparing to the current SOTA task on both class and task incremental learning methods.
 \begin{figure*}[tb]
	\centering
	\includegraphics[clip, trim=0.0cm 12cm 0cm 0cm, width=1\textwidth]{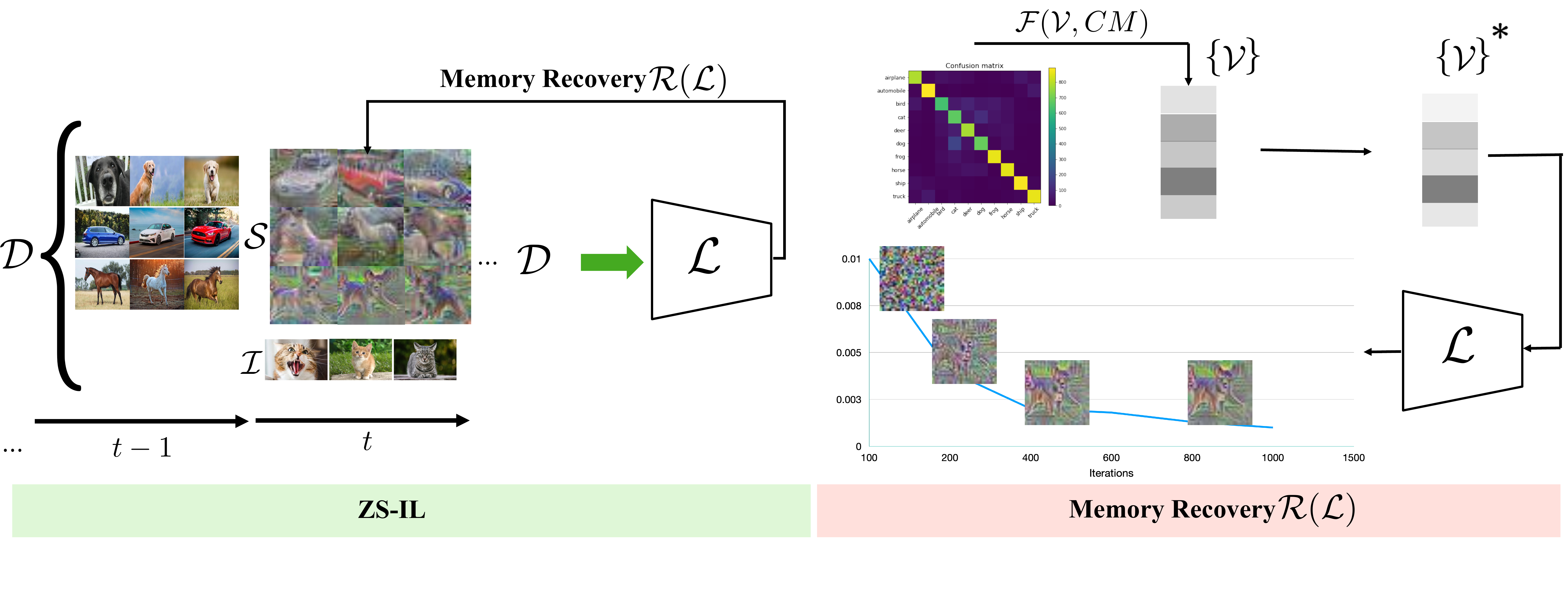}
	\vspace{-9mm}
	\caption {Overview of the proposed approach. When a new incremented class(es) $I$ emerge, the memory recovery paradigm $\mathcal{R}(\mathcal{L})$ is called to generate transfer set $\mathcal{S}$, including synthesized images for the classes that the learner network has learned before. In the proposed recovery engine, we first create model outputs for each class $\mathcal{\{V\}}^{*}$ in two steps. Then, a random noise input is initialized to be optimized conditioned on the generated output. This procedure is performed for each class several times to form the transfer set $\mathcal{S}$. Finally, the learner network is retrained on the combined dataset $\mathcal{D}$, including the transfer set $\mathcal{S}$ and new incremented class $\mathcal{I}$ using a two-part cost function composing a classification (CE) and knowledge distillation (KD) terms, respectively.}
	\label{fig:proposed_method}
\end{figure*}
\vspace{-2mm}
\section{Related Work}
\vspace{-2mm}
\textbf{Incremental Learning} Catastrophic forgetting has been a well-known problem of artificial neural networks where the model parameters are overwritten when retraining the model with new data. Several studies have been carried out to tackle this problem which falls into four categories as follows.

\textit{Model Growing.} Model growing methods are the first investigations into incremental learning in which a new set of network parameters is dedicated for each task in a dynamic architecture networks\cite{aljundi2017expert,xu2018reinforced,yoon2017lifelong,rosenfeld2018incremental}. Therefore, they require the task label to be known at test time to trigger the correct state of the network, which can not be employed in the class incremental setting where task labels are not provided. Moreover, maintaining class incremental performance can be met at the cost of limited scalability.
\textit{Regularization-based.} The regularization-based approaches penalize the model to update parameters that are critically important for earlier tasks by adding a particular regularization term to the loss function. The importance score of each parameter can be estimated through the Fisher Information Matrix\cite{kirkpatrick2017overcoming}, gradient extent\cite{aljundi2018memory}, the path integral formulation for the loss gradients\cite{zenke2017continual}, and uncertainty estimation using Bayesian Neural Network\cite{nguyen2017variational}. Despite the fact that the vast amount of literature falls into this family,  the suggested soft penalties might not be sufficient to restrict the optimization process to stay in the feasible region of previous tasks, particularly with long sequences\cite{farquhar2018towards}, which caused an increased forgetting of earlier tasks.
\textit{Memory Replay} memory replay methods allow access to a fixed memory buffer to hold actual sample \cite{rebuffi2017icarl,isele2018selective,chaudhry2019continual,rolnick2018experience} data or pseudo-samples generated by a generative model\cite{robins1995catastrophic,atkinson1802pseudo} from former tasks. When learning new tasks, samples from the former tasks are either reused as model inputs for rehearsal or to constrain optimization\cite{lopez2017gradient,chaudhry2018efficient} of the recent task loss to prevent preceding task interference. One another use of memory buffer is using dark knowledge\cite{hinton2015distilling} by utilizing the model's past parametrizations as a teacher and transferring knowledge to a new model while learning new tasks. One typical usage is incorporating the distillation loss on past exemplars with the classification loss for learning from new class training samples\cite{li2017learning}. Although this approach is pioneering, its main limitation is a fixed budget, which caused the bias problem\cite{hou2019learning,wu2019large} by the imbalanced number of old and new class training samples, which destroys the performance\cite{belouadah2019il2m}.
\textit{Parameter Isolation.} This group of methods relies on estimating binary masks that allocate a fixed number of parameters for each task where the network architecture remains static. Such masks can be estimated either by random assignment\cite{masse2018alleviating}, pruning\cite{mallya2018packnet}, or gradient descent\cite{mallya2018piggyback,serra2018overcoming}. However, this family is limited to the task incremental setting and better agreed for learning a long sequence of tasks when model capacity is not a concern, and optimal performance is the priority.

\textbf{Knowledge Distillation}
Knowledge Distillation (KD) is first proposed by Hinton et al. \cite{hinton2015distilling} to transfer knowledge from a teacher classifier to a student classifier. This subject typically has been attended for almost two reasons. (1) Sharing knowledge from a substantial deep neural network, named as a teacher, with a high learning power to a more compact network with the purpose of compressing the network structure\cite{polino2018model,vapnik2015learning}. (2) Distillation the teacher knowledge to an arbitrary student network where the used training data are not available at the current time\cite{nayak2019zero}. Compared with these reasons, the second one has least investigated in the KD literature where is the focus of our work. In this case, the pre-trained network is taken as the teacher, and the same network is employed as the student to adapt to new classes while distilling knowledge on the former classes from the teacher network.
\vspace{-2mm}
\section{Problem setting}
\vspace{-2mm}
Without loss of generality, a classical IL scenario can be assumed as a set \(T=\left\{T_{1}, \ldots, T_{N}\right\}\) of \(N\) different tasks. Each task \(T_{i}\) takes the form of a classification problem, \(T_{i}=\) \(\left\{\mathbf{x}_{j}, y_{j}\right\}_{j=1}^{N \times l},\) where \(y_{j} \in\left\{1, \ldots, l\right\}\). When a new task \(T_{t+1}\) arrives we ask the network to learn to classify the new images according to \(y_{t+1}\), being simultaneously able to solve the former tasks. Our method only relies on a single DNN network, named the learner network $\mathcal{L}$, in all steps with no external memory and no parallel network.

Following common practice in IL literature, we formalize the setting conditioned on the absence of a task oracle at test time as follows: Task-IL and Class-IL. In the Task-IL setting, the learner network $\mathcal{L}$ is composed of a feature extractor followed by a multiple heads where each head is composed of a linear classifier with a softmax activation which computes task-specific classification probabilities. On the contrary, in the Class-IL setting, the learner network has a single head while is shared over all the tasks that make the Class-IL more challenging. In this paper, we consider both settings and report the results regarding them.
\vspace{-7mm}
\section{Proposed method}
\vspace{-2mm}
Availability of the actual data samples can enable solving for straightforward retraining, thus learning incrementally that can overcome catastrophic forgetting problem. However, preserving earlier data is not straightforward in real-world scenarios due to computational overhead and privacy concerns. In this paper, the main idea is to recall (instead of preserving) past exemplars, known as network experiences, without using either a parallel network or memory buffer. To this end, we propose ZS-IL in which, whenever a new task arises, we trigger a memory recovery paradigm $\mathcal{R}(\mathcal{L})$ to obtain a transfer set $\mathcal{S}$ of past experiences as surrogates for the earlier tasks observations. Then, we form a combined training dataset $\mathcal{D}$ from the recovered transfer set $\mathcal{S}$ and the current incremented class(es) $\mathcal{I}$ data. Thus, the learner network $\mathcal{L}$, which is supposed to learn the task sequences \(T\) incrementally, is retrained on  $\mathcal{D}$ in each time step. In the suggested memory recovery paradigm, we first model the learner network output $\mathcal{L}$ regarding the case when a real sample from the past is fed to the model. These outputs are provided by generating initial output vectors $\{\mathcal{V}\}$ and refining them by applying the $\mathcal{F}(\mathcal{V}, CM)$ constraint, as a supervision, to perform adjustment due to underlying distribution obtained from a dynamic Confusion Matrix ($CM$), resulting in reaching $\{\mathcal{V}\}^\ast$. Fig.~\ref{fig:proposed_method} shows a clear overview of the proposed method. In the rest of this section, we first describe the proposed ZS-IL and then outline the suggested memory recovery paradigm in detail.

\begin{algorithm}[t]
\DontPrintSemicolon
\SetKwInOut{Input}{Input}
\SetAlgoLined
\SetKwInOut{Require}{Require}
\SetNoFillComment
\LinesNotNumbered 
 \Input{Task sequence set $T$, Transfer set size $K$, Learner model parameters $\theta_{\mathcal{L}}$}
 \Require{Confusion Matrix $CM$ from $\mathcal{L}$, Incremented class(es) $\mathcal{I}$}
 \tcp{memory recovery paradigm}
 $\mathcal{S}\gets$ $\mathcal{R}(\theta_L, K, CM)$\;
  $L:$ number of samples in $\mathcal{I}$ \;
   \tcp{store network outputs and Ground Truth for $\mathcal{S}$ with pre-update parameters}
 \For{$(x,y)$ in $\mathcal{S}$}{
    $\mathcal{O} \gets \mathcal{L}(x; \theta_{\mathcal{L}})$ \;
  }
 \tcp{form combined training set}
 $\mathcal{D} \gets \bigcup_{i=1}^{i=L}\{(x^i, y^i, -): x^i \in \mathcal{I}_t\} \quad \cup$ \; $\quad\quad\quad\quad\quad\quad \bigcup_{j=1}^{j=K}\{(x^j, y^j, \mathcal{\mathcal{O}}^j): x^j \in \mathcal{S}_t\}$\;
  \tcp{define a two-parted loss function}
  $\ell_{Total}=\underset{(x,y,\mathcal{O})\in\mathcal{D}}{\Sigma}[\Sigma_{i=1}^N \ell_{CE}(\mathcal{L}(x^i; \theta_{\mathcal{L}}),y^i)] \quad + $ \;   $\quad\quad\quad\quad\quad\quad\quad\quad\quad \lambda[\Sigma_{j=1}^K \ell_{KD}(\mathcal{L}(x^j;\theta_{\mathcal{L}}),\mathcal{O}^j)] $ \;
  \tcp{optimize the network for the total loss}
   \For{$(x,y)$ in $range(K+L)$}{ 
        sample a batch $b_1$ form $\mathcal{D}$, where $(x,y,\mathcal{O}) \in \mathcal{S} $\;
        sample a batch $b_2$ form $\mathcal{D}$, where $(x,y) \in \mathcal{I} $\;
        update $\theta_{\mathcal{L}}$ by taking a SGD step on $b_1+b_2$ loss $\ell_{Total}(\theta_{\mathcal{L}})$

  }
  
 \caption{Zero-Shot Incremental Learning}
  \label{fig:ZSIL}
\end{algorithm}
\begin{algorithm}[t]
\DontPrintSemicolon
\SetAlgoLined
\SetNoFillComment
\LinesNotNumbered 
\SetKwInOut{Input}{Input}
\SetKwInOut{Output}{Output}
\SetKwInOut{Require}{Require}
\SetKwRepeat{Do}{do}{while}
 \Input{Learner model parameters $\theta_{\mathcal{L}}$, Transfer set size $K$, Confusion Matrix $CM$}
 \Output{Transfer set $\mathcal{S}$} 
 \Require{Dirichlet distribution parameters $\alpha, \beta$, Thresh. $\eta$, Temperature for distillation $\tau$ }\;
 $k:$ number of classes from $\mathcal{L}$ \;
 $p:$ $K/k$  \tcp*{Number of samples per class}
 $\mathcal{S} \gets \{ \}$ \;
 Constraint $\mathcal{F}(\mathcal{X}, \gamma) \gets (\left\|\mathcal{X} - \gamma\right\|_{2}^{2} < \eta) $\;
  \For{$i=1:k$}{
      \For{$j=1:p$}{
      \Do{Constraint $\mathcal{F}(\mathcal{V}^i_j, CM^i)$ passed}{ $\text { Sample } \mathcal{V}_{j}^{i} \gets \operatorname{Dir}\left(d, \beta \times \boldsymbol{\alpha}^{i}\right)$ \; $\mathcal{V^\ast}^{i}_{j} \gets \mathcal{V}^i_j $}
        Initialize $x^i_j$ to random noise\;
        $\bar{x}^i_j \gets$ optimize $x^i_j$ with \; $\quad\quad\quad\quad \ell_{CE}\left(\mathcal{V^\ast}_{j}^{i}, \mathcal{L}\left(x^i_j, \theta_{\mathcal{L}}, \tau \right)\right)$\;
         $\mathcal{S} \gets \mathcal{S} \cup \bar{x}^i_j$ \;
      }
  }
  
  $\mathcal{S} \gets Augmentation(\mathcal{S}) $ 

 \caption{Memory Recovery Paradigm}
 \label{fig:MR}
\end{algorithm}
\subsection{Zero-Shot Incremental Learning} In this work, the zero-shot incremental learning term expresses that there is no need for a memory buffer to learn incrementally, while the DNN model complexity is fixed. Besides, we synthesize a transfer set \(\mathcal{S}\) as past experiences of the network in a zero-shot manner. To put this idea into practice, we run the following steps, which are listed in Alg.~ \ref{fig:ZSIL}. Whenever ZS-IL receives a new incremented class(es) data $\mathcal{I}= \left\{\left(x_{i}, y_{i}\right), 1 \leq i \leq L, y_{i} \in[1, \ldots, l]\right\} $, where $L$ is the total number samples of $l$ classes belong to the current task $T_t$, it initiates the memory recovery procedure $\mathcal{R}(\mathcal{L})$ (Alg.~\ref{fig:MR}) to create the transfer set \(\mathcal{S} =\left\{\left(\hat{x}_{j}, \hat{y}_{j}\right), 1 \leq j \leq K, \hat{y}_{j} \in\right. [1, \ldots, k]\}  \), where $K$ is the total number of synthesized samples of $k$ classes that have been learned by the learner network $\mathcal{L}$. Next, the samples from the transfer set \(\mathcal{S}\) is fed to the learner network, and the resulting network’s logits \(\mathcal{O}=\left[o_{1}(x), \ldots, o_{K}(x)\right]\) for all samples are stored. Then, we form a combined dataset $\mathcal{D}$ per each task $T_t, t \in [1, \ldots, N]$ in the task sequence $T$. Finally, the learner network \(\mathcal{L}\) parameters are updated to minimize a cost function such a way that each data sample from the new incremented class $\mathcal{I}$ will be classified correctly, as classification loss, and for the samples in $\mathcal{S}$, current network’s logits will be reproduced as close as those have been stored in the previous step, as knowledge distillation loss by KD term. 

We employ the softmax cross entropy as the classification loss, which is computed using Equ. \ref{eq:eq1}:

\begin{equation}
\ell_{CE}(\mathcal{L}(x; \theta_{\mathcal{L}}),y)=- \sum_{(x,y) \in \mathcal{I}} \sum_{i=1}^{l}\delta_{y=i} \log \left[p_{i}(x)\right]
\label{eq:eq1}
\end{equation}
where $\delta_{y=i}$ is the indicator function and $p_i(x)$ is the output probability (i.e. softmax of logits) of the $i^{th}$ class in $l$ new incremented class(es).

For the distillation purpose, we adopt knowledge distillation from network output (logits), know as dark knowledge\cite{hinton2015distilling,wang2020knowledge,buzzega2020dark}, and our objective is to minimize the Euclidean distance between the stored logits in the transfer set \(\mathcal{O}=\left[o_{1}(x), \ldots, o_{K}(x)\right]\) and those generated by the learner network \(\hat{\mathcal{O}}=\left[\hat{o}_{1}(x), \ldots, \hat{o}_{K}(x)\right]\) as follows:
\begin{equation}
\ell_{KD}(\mathcal{L}(x;\theta_{\mathcal{L}}), \mathcal{O})=\sum_{(x,y) \in \mathcal{S}} \sum_{j=1}^{k} \left\|o^j(x)-\hat{o}^j(x)\right\|_{2}^{2}
\end{equation}
Having a look at both cost functions, we define the total loss function by a linear combination as bellow:
\begin{equation}
\ell_{Total} = \ell_{CE} + \lambda \ell_{KD}
\label{eq:totalLoss}
\end{equation}
where $\lambda$ is predefined parameter to control the degree of distillation.
\subsection{Memory Recovery Paradigm}
 The memory recovery paradigm is devised to synthesis the previously learned experiences (\ie knowledge) by $\mathcal{L}$. In particular, we synthesis past samples in a way that the learner network strongly believes them to be actual samples that belong to categories in the underlying data distribution. Therefore, they are the network acquaintance that might not be natural-looking data.
 It is worth noting that we only use the learned parameters of the learner model $\mathcal{L}$ since they can be interpreted as the memory of the model in which the essence of training has been saved and encoded. 

In the suggested paradigm, we first model the network output space using a two-stepped approach and then generate synthesized samples by back-propagating the modeled output through the network, where each is described in the rest.The whole procedure is shown in Alg.~\ref{fig:MR}.

\textbf{Modeling the Network Output Space} In the first phase, we model the output space of the learner network. Suppose \(y^{i,j} \sim \mathcal{L}(x, \theta_{\mathcal{L}})^{(T_i, C_j)} \) is the actual model output when \(j^{th}\) class of \(i^{th}\) task observation was fed to the network. To reproduce such output in the absence of actual previous one, $(\mathcal{V^\ast})^{i,j} \sim y^{i,j}$, we consider a two-stepped approach as is explained bellow.

In ~\cite{nayak2019zero}, it is investigated that sampling  the output  vector $\VV$  from a  Dirichlet Distribution is a straightforward way for modeling the output of the learner network. The distribution of vectors that their ingredients are in the range $[0, 1]$ and whose sum is one as detailed in \cite{nayak2019zero}. Despite interesting results, this method lacks preserving extra class similarity, hence generating outlier vectors that do not follow the networks underlying distribution. Thus, false memory happens when we query the memory for a specific target. In this case, the result of memory recovery is a mixture of some targets that do not really exist in the past experienced, resulting in confusing the learner in the retraining phase. In our novel memory recovery process, we make supervision as the second step, to detect such outliers to boost the retrieved memory in terms of generating a single-target-based transfer set. To do this, we make supervision by applying a constraint on the generated vector $\mathcal{F}(\mathcal{V}, \gamma)$, in which an arbitrary generated vector from the previous step $\mathcal{V}$  is a good candidate only if it has a distance lower than a predefined threshold $\eta$ to a dynamic recommender $\gamma$. This recommendation vector supposed to be declared at a low-cost and represent the class similarity as well. Considering these concerns, we imply a dynamic confusion matrix $CM$ that is constructed incrementally when training each task. An arbitrary confusion matrix is a table that is typically used to describe the performance of a classification model. Intuitively, when a model wrongly classifies some samples to class \(c\), it means class \(c\) is highly correlated with the target class. On the other side, when a model always correctly classifies class \(c\) from the target class, it means class \(c\) is highly distinguishable from the target class. This is the exact expected result from a DNN model output. Mathematically, we check if the generated vector $\mathcal{V}^i$ for arbitrary class $i$,  has a distance smaller than a predefined threshold $\eta$ to the $i^{th}$ row of the constructed dynamic confusion matrix $CM^i$, so as to obtain the checked output vector $\mathcal{V^\ast}^i$.

\textbf{Back propagate Through the Network} In the second phase, conditioned on the network's generated outputs $\{\mathcal{V^\ast}\}$, we synthesize a transfer set \(\mathcal{S}\). For an arbitrary generated vector $\mathcal{V}^{\ast}$, we start with a random noisy image $x$ sampled from a uniform distribution in the range $[0, 1]$ and update it till the cross-entropy (CE) loss between the generated model output \(\mathcal{V}\ast\) and the model output predicted by the learner network \(\mathcal{L}\left(x, \theta_{T}, \tau\right) \) is minimized.
\begin{equation}
\bar{x}=\operatorname{argmin} \ell_{CE}\left(\mathcal{L}\left(x, \theta_{\mathcal{L}}, \tau\right) , \boldsymbol{y}\right)
\end{equation}
Where \(\tau\) is the temperature value used in the softmax layer\cite{hinton2015distilling} for the distillation purpose. This procedure is renewed for each of the \(k\) classes has been learned $K/k$ times, where $K$ is the whole number of samples to be formed the transfer set $\mathcal{S}$. Moreover, to have diverse transfer set, we perform typical data augmentations such as random rotation in $[-50,50]$, scaling by a factor randomly selected from \(\{0.95, 0.975, 1.0, 1.025\}\), RGB jittering, and random cropping. Along with the above typical augmentations, we also add random uniform noise in the range $[-10,10]$. This augmentation aims to synthesize a robust transfer set that behaves similarly to natural samples regarding the augmentations and random noise. 

The proposed paradigm is a remedy for memory bottlenecks so that the model alone can alleviate catastrophic forgetting, resulting in remarkable performance among the data-free works on both task-IL and class-IL, as will be seen in the next section.
 \begin{figure*}[tb]
	\centering
	\vspace{-3mm}
\subfloat{\label{figur:1}\includegraphics[width=16mm]{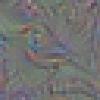}} \vspace{0.2mm}
\subfloat{\label{figur:1}\includegraphics[width=16mm]{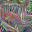}} \vspace{0.2mm}
\subfloat{\label{figur:1}\includegraphics[width=16mm]{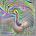}} \vspace{0.2mm}
\subfloat{\label{figur:1}\includegraphics[width=16mm]{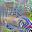}} \vspace{0.2mm}
\subfloat{\label{figur:1}\includegraphics[width=16mm]{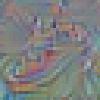}} \vspace{0.2mm}
\subfloat{\label{figur:1}\includegraphics[width=16mm]{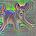}} \vspace{0.2mm}
\subfloat{\label{figur:1}\includegraphics[width=16mm]{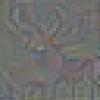}} \vspace{0.2mm}
\subfloat{\label{figur:1}\includegraphics[width=16mm]{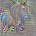}} \vspace{0.2mm}
\subfloat{\label{figur:1}\includegraphics[width=16mm]{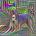}} \vspace{0.2mm}
\subfloat{\label{figur:1}\includegraphics[width=16mm]{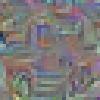}} 
\vspace{-2mm}
	\caption{Visualizing sampled recovered data from the learner network using the proposed memory recovery paradigm, while training on CIFAR10 dataset.}
	\label{fig:impressions}
\end{figure*}
\vspace{-2mm}
\section{Experiments}
\vspace{-2mm}
This section demonstrates the empirical effectiveness of our proposed method in various settings over two benchmark classification datasets compared with recent SOTA works. We also perform ablation experiments to further analyze different components of our method.

\subsection{Implementation Details}
\textbf{ZS-IL.} For the learner network architecture, we follow \cite{rebuffi2017icarl} and use ResNet18\cite{he2016deep} without pre-training. To perform a fair comparison with other IL methods, we train the networks using the Stochastic Gradient Descent (SGD) optimizer with a learning rate of $0.1$ with other parameters set to their default values. We consider the number of epochs per task concerning the dataset complexity; thus, we set it to 50 for Sequential CIFAR-10 and 100 for Sequential Tiny-ImageNet, respectively, which is commonly made by prior works\cite{rebuffi2017icarl,wu2019large,zenke2017continual}. In training duration, we select a batch of data from the current task and a minibatch of data from the transfer set \(\mathcal{S} \), where depending on the hardware restriction, we set both to $32$.

\textbf{Memory Recovery Paradigm.} We consider the temperature value $\tau$ to $20$ for the distillation purpose. In the Dirichlet distribution, we set $\beta$ in $[1,0.1]$ for each dataset, where half the transfer set $\mathcal{S}$ is synthesized by $1$ and the rest with $0.1$ as in \cite{nayak2019zero}. The transfer set size is set to $6k$ (the effect of transfer size is explored in Section \ref{ablation_study}). The $\eta$ value in the constraint step is empirically valued at $0.7$. To optimize the random noisy image, we employ the Adam optimizer with a learning rate of $0.01$, while the maximum number of iterations is set to $1500$.
\vspace{-2mm}
\subsection{Evaluation Metric}
\vspace{-2mm}
For all experiments, we consider the average accuracy across tasks as the evaluation criteria, following previous works\cite{buzzega2020dark,rebuffi2017icarl}. Average accuracy \((A \in[0,100])\) after learning the \(k^{t h} \operatorname{task}\left(T_{k}\right)\), is defined as \(A_{k}=\frac{1}{k} \sum_{j=1}^{k} a_{k, j} ;\) where \(a_{k, j}\) is the performance of the model on the test set of task $j$, after the model is trained on task $k$.
\vspace{-2mm}
\subsection{Datasets}
\vspace{-2mm}
 Two commonly-used public datasets is exploited  for evaluating the proposed method:(1) CIFAR-10\cite{krizhevsky2009learning} and (2)Tiny-ImageNet\cite{le2015tiny}.
 
\textbf{Tiny-ImageNet} dataset is a subset of 200 classes from ImageNet\cite{deng2009imagenet}, rescaled to image size $64\times64$, where each class holds $500$ samples. In order to form a balanced dataset, we select an equal amount of $20$ randomly chosen classes for each task in a sequence of $10$ sequential tasks.  

\textbf{CIFAR-10} contains $60000$ labeled images, split into $10$ classes, roughly $6k$ images per class. All of the images are small and have $32\times32$ pixels. On this dataset, we choose 2 random classes for each task, resulting in $5$ consecutive tasks.
\subsection{Comparing Methods}
Despite the fact that recent SOTA performance has been achieved by those methods that buffer past samples, an increasing number of researches have been investigated on the data-free approaches since the limitations in storing the former data as mentioned before. Our proposed ZS-IL falls into data-free categories since it only relies on the learner network with no additional equipment. So, our method has a significant advantage despite memory-based works.

Regarding prominent works in the data-free group, we evaluate our method against (1) regularization-based works oEWC\cite{schwarz2018progress}, SI\cite{zenke2017continual}, ALASSO\cite{park2019continual}, UCB\cite{ebrahimi2019uncertainty} (2) knowledge distillation based method LwF\cite{li2017learning}, (3) dynamic architecture approach PNN\cite{rusu2016progressive}, (4) Pseudo-rehearsal models, where past samples are generated with generative models DGM\cite{ostapenko2019learning}, DGR\cite{shin2017continual}, MeRGAN.

To further examine the proposed ZS-IL, we make also a comparison with SOTA memory-replay methods. In this category, stored data are either reused as model inputs for rehearsal purposes MER\cite{riemer2018learning}, iCaRL\cite{rebuffi2017icarl}, DER\cite{buzzega2020dark}, or to constrain optimization of the new task loss to prevent previous task interference GEM\cite{lopez2017gradient},a-GEM\cite{chaudhry2018efficient}, GSS\cite{aljundi2019gradient}, TPCIL\cite{tao2020topology}, Mnemonics \cite{liu2020mnemonics}.
\subsection{Results}
\textbf{Data-free works:} Table~\ref{tab:resultDataFree} shows the performance of  ZS-IL in comparison to the other mentioned methods on the two commonly used considered datasets. Our proposed method \ie ZS-IL has achieved the SOTA performance in almost all settings. Compared with regularization-based methods \ie SI\cite{zenke2017continual}, oEWC\cite{schwarz2018progress}, and ALASSO\cite{park2019continual} the gap does not seem to be bridged, indicating that the old parameter set's regularization is not sufficient to mitigate forgetting since the wights' importance, calculated in an earlier task, might be unreliable in later ones. While remaining computationally more effective, LWF\cite{li2017learning} as a KD based method shows worse than SI\cite{zenke2017continual} and oEWC\cite{schwarz2018progress} on average. PNN\cite{rusu2016progressive}, one of the dynamic architecture works, produces the strongest results in the task-IL setting between data-free methods, specifically $95.13$\% and $67.84\%$ in CIFAR-10 and tiny-ImageNet, respectively. However, it suffers from an exponential memory increasing issue, making it impracticable in the class-IL setting and challenging datasets with more classes. Besides, our method won the second place in the Task-IL setting at the cost of keeping the network architecture fixed with margins $2.01$\% on CIFAR-10 and $0.42$\% on Tiny-ImageNet than the top rank.Moreover, a comparison between our method and pseudo-rehearsal-based methods like DGM\cite{ostapenko2019learning}, DGR\cite{shin2017continual}, and MeRGAN\cite{wu2018memory}, which have recently gained a lot of attention due to the power of simulating past experiences than storing them, further verifies the effectiveness of our method. In particular, our method achieves a relative improvement of $3.40$\%,$3.21$\%,$4.61$\%,$6.1$\% over DGM\cite{ostapenko2019learning}.

\textbf{Memory-based works:} Table~\ref{tab:resultMemReplay} illustrates the contrastive performance comparison results against memory-based works. All methods in the table are similar in terms of buffering past samples regardless of their training approach. For a fair comparison among them, we report results, while the maximum size to buffer real past samples is $500$. Besides, our proposed method is not allowed to store any samples from the past. Even though that there is a significant difference, our method can achieve excellent results in both class-IL and task-IL on both datasets. In particular, our proposed method obtains $75.43$\%, $93.12$\% in the class-IL, Task-IL settings on the CIFAR-10, and $733.06$\%, $67.42$\% in the class-IL, Task-IL settings on the Tiny-ImageNet, respectively, making the relative improvement over the recent SOTA TPCIL\cite{tao2020topology} by $16.62$\%, $8.03$\% in the class-IL settings on the CIFAR-10 and Tiny-ImageNet, and $19.01$\% in the task-IL in the more challenging Tiny-ImageNet respectively.

\begin{table}[t]
\begin{center}
\caption{\footnotesize{Classification results (average accuracy) for standard data-free IL benchmarks. "-" denotes experiments we were unable to run, because of compatibility issues. The largest number in each column is marked in bold-face and the subscript of each metric indicates top-2 ranking.}}
\vspace{-2mm}
\label{tab:resultDataFree}  
\begin{tabular*}{0.5\textwidth}{l@{\extracolsep{\fill}}cc cc}
\hline
\multicolumn{1}{l}{\multirow{2}{*}{Method}} & \multicolumn{2}{c}{CIFAR-10} & \multicolumn{2}{c}{Tiny-ImageNet} \\ 
\multicolumn{1}{c}{} & Class-IL & Task-IL & Class-IL & Task-IL \\ \hline
JOINT & 92.02 & 98.29 & 59.45 & 82.07 \\ \hline
oEWC\cite{schwarz2018progress} & 19.49 & 68.29 & 7.58 & 19.20 \\
SI\cite{zenke2017continual} & 19.48 & 68.05 & 6.58 & 36.32 \\
ALASSO\cite{park2019continual} & 25.19 & 73.79 & 17.02 & 48.07 \\
LwF\cite{li2017learning} & 19.61 & 63.29 & 8.46 & 15.85 \\
PNN\cite{rusu2016progressive} & - & \textbf{95.13}\textsuperscript{\textbf{1}} & - & \textbf{67.84}\textsuperscript{1} \\
UCB\cite{ebrahimi2019uncertainty} & 56.23 & 78.56 & 23.43 & 49.01 \\
DGM\cite{ostapenko2019learning} & \textbf{71.94}\textsuperscript{\textbf{2}} & 89.91 & \textbf{28.45}\textsuperscript{\textbf{2}} & 61.32 \\
DGR\cite{shin2017continual} & 49.69 & 79.86 & 17.38 & 38.41 \\
MeRGAN\cite{wu2018memory} & 66.76 & 84.76 & 23.86 & 58.32 \\ \hline
\textbf{ZS-IL(Ours)} & \textbf{75.34}\textsuperscript{\textbf{1}} & \textbf{93.12}\textsuperscript{\textbf{2}} & \textbf{33.06}\textsuperscript{\textbf{1}} & \textbf{67.42}\textsuperscript{\textbf{2}} \\ \hline
\end{tabular*}
\end{center}
\end{table}
\begin{table}[t]
\begin{center}
\caption{\footnotesize{Classification results (average accuracy) for standard memory-replay based IL benchmarks on the well-known visual datasets. "-" indicates experiments have unmanageable training time issues. The largest number in each column is marked in bold-face and the subscript of each metric indicates top-2 ranking.}} \vspace{-2mm}
\label{tab:resultMemReplay}  
\begin{tabular*}{0.5\textwidth}{l@{\extracolsep{\fill}}cc cc}
\hline
\multicolumn{1}{l}{\multirow{2}{*}{Method}} & \multicolumn{2}{c}{CIFAR-10} & \multicolumn{2}{c}{Tiny-ImageNet} \\ 
\multicolumn{1}{c}{} & Class-IL & Task-IL & Class-IL & Task-IL \\ \hline
JOINT & 92.02 & 98.29 & 59.45 & 82.07 \\ \hline
MER\cite{riemer2018learning} & 57.74 & 93.61 & 9.99 & 48.64 \\
GEM\cite{lopez2017gradient} & 26.20 & 92.16 & - & - \\
a-GEM\cite{chaudhry2018efficient} & 22.67 & 89.48 & 8.06 & 25.33 \\
iCaRL\cite{rebuffi2017icarl} & 47.55 & 88.22 & 9.38 & 31.55 \\
FDR\cite{benjamin2018measuring} & 28.71 & 93.29 & 10.54 & 49.88 \\
GSS\cite{aljundi2019gradient} & 49.73 & 91.02 & - & - \\
HAL\cite{chaudhry2020using} & 41.79 & 84.54 & - & - \\
DER\cite{buzzega2020dark} & \textbf{70.51}\textsuperscript{\textbf{2}} & 93.40 & 17.75 & 51.78 \\
BIC\cite{wu2019large} & 49.18 & 89.91 & 11.09 & 32.94 \\
WA\cite{zhao2020maintaining} & 53.76 & 94.21 & 15.42 & 37.12 \\
Mnemonics \cite{liu2020mnemonics} & 57.26 & \textbf{95.91}\textsuperscript{\textbf{2}} & 23.47 & 42.36 \\
TPCIL\cite{tao2020topology} & 58.72 & \textbf{96.67}\textsuperscript{\textbf{1}} & \textbf{25.03}\textsuperscript{\textbf{2}} &  \textbf{48.41}\textsuperscript{\textbf{2}} \\ \hline
\textbf{ZS-IL(Ours)} & \textbf{75.34}\textsuperscript{\textbf{1}} & 93.12 & \textbf{33.06}\textsuperscript{\textbf{1}} & \textbf{67.42}\textsuperscript{\textbf{2}} \\ \hline
\end{tabular*}
\end{center}
\end{table}
\vspace{-2mm}
    \begin{figure}[!htp]
        \centering
        \subfloat{\label{figur:1}\includegraphics[clip, trim=0cm 20cm 48cm 0cm, width=0.22\textwidth]{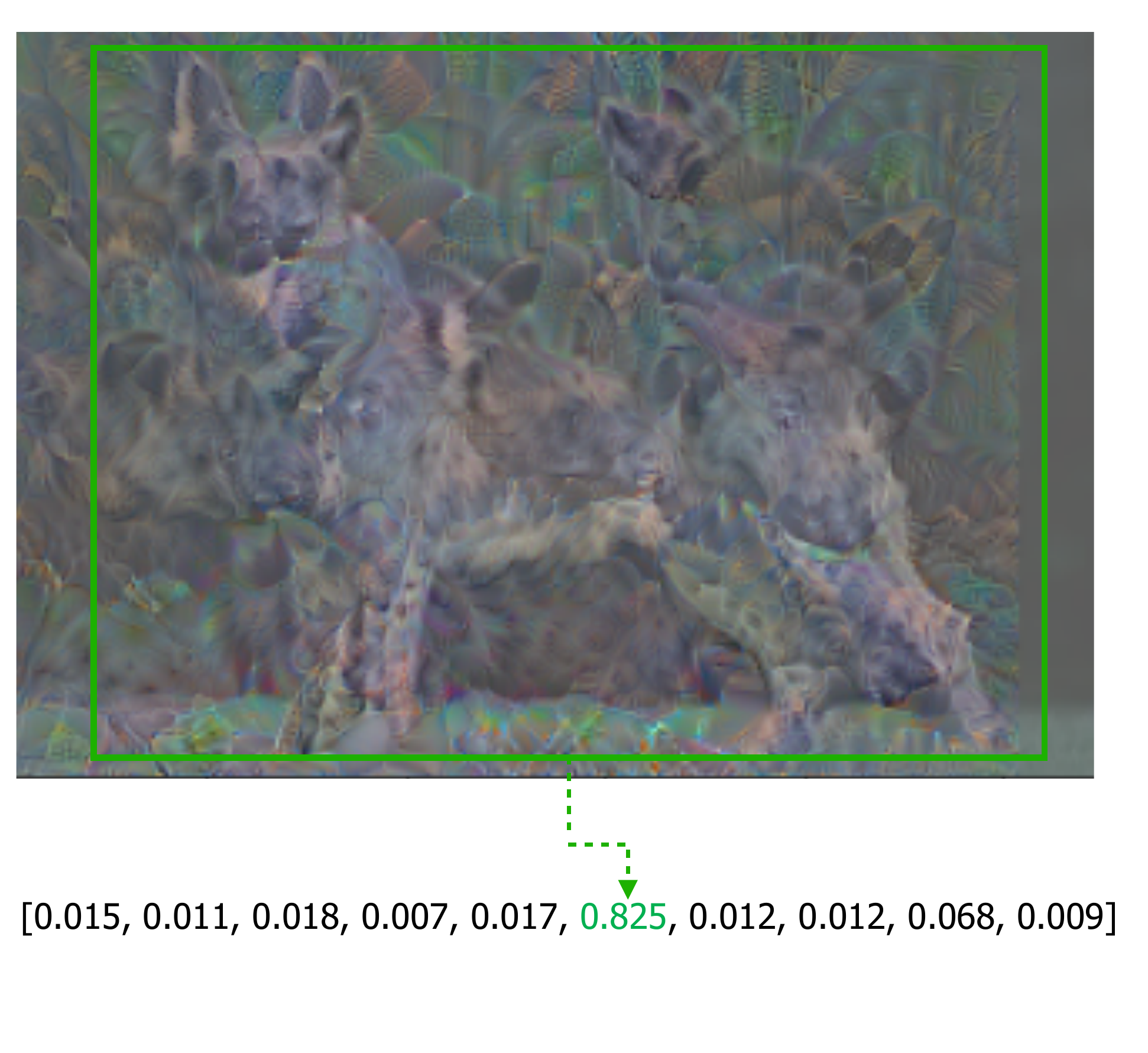}}
        \subfloat{\label{figur:1}\includegraphics[clip, trim=0cm 20cm 48cm 0cm, width=0.22\textwidth]{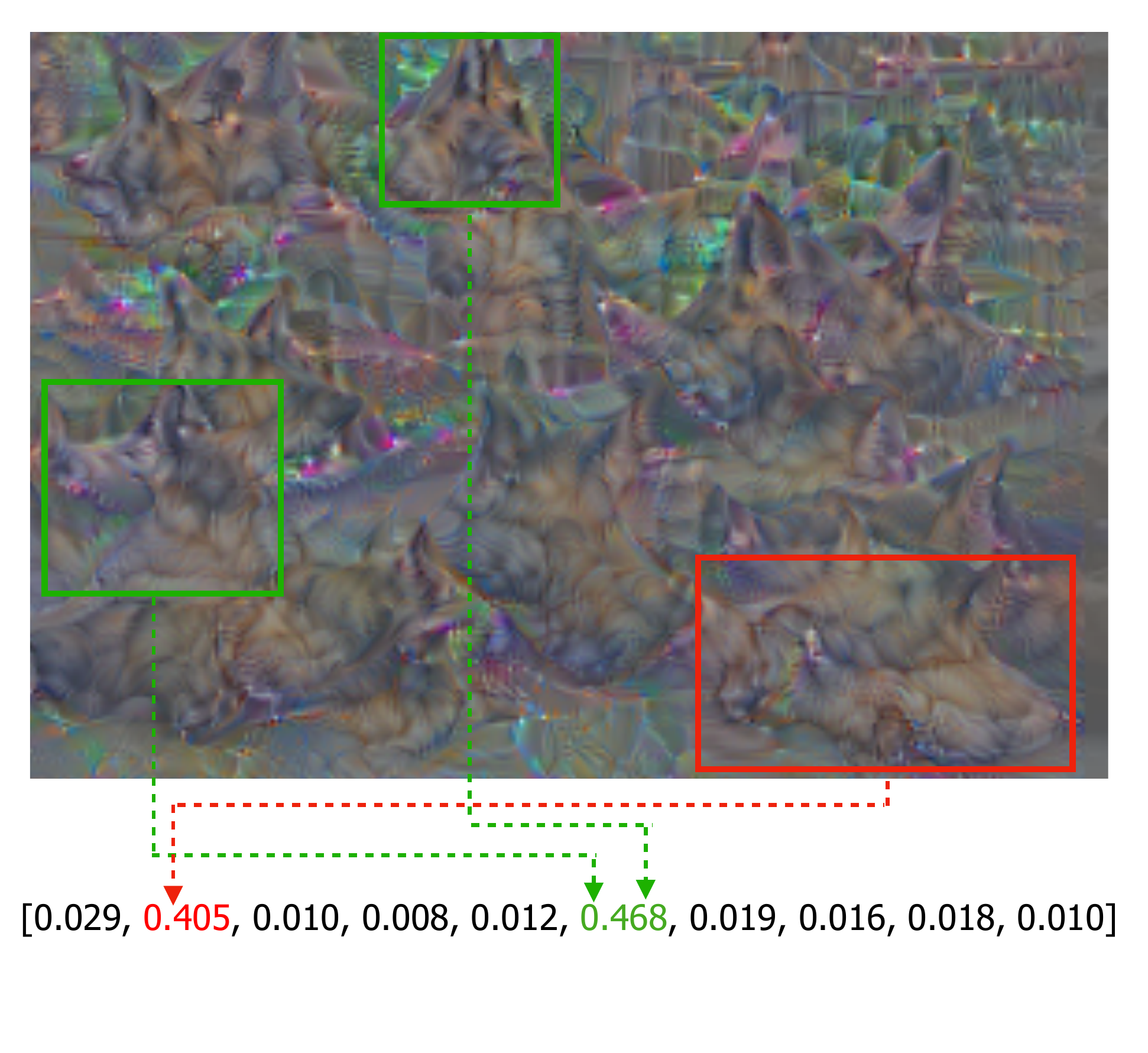}} 
        \caption{Visualization of happened False Memory when retrieving data from the memory recovery step (left) and a correct sample which passed the second step in our introduced network output modeling procedure (right). }
	\label{fig:FM}
    \end{figure}
       
\subsection{Ablation Studies}\label{ablation_study}
\vspace{-1mm}
\textbf{Synthesized Transfer Set} As discussed before, we follow a two-stepped strategy to generate the network output in our proposed model memory recovery process. Firstly, we follow the approach adopted in \cite{nayak2019zero} in which a Dirichlet distribution $\operatorname{Dir}(d, \beta \times \boldsymbol{\alpha})$ is utilized to generate the vectors in sotmax space, where we need to set two parameters in this distribution: $\alpha$, and $\beta$ ($d$ is the dimension of the expected vector) . $\alpha$ is the concentration parameter to control the probability mass of a sample from the distribution. Decreasing its value caused a vector that is hugely concentrated in only a few ingredients and vice versa. $\alpha$ is interpreted as a class similarity vector obtained from the network's final layer and can be scaled with the $\beta$ parameter.

As mentioned before, this method encounters false memory since some generated vectors are a mix-up of several targets. To tackle this problem, we suggest applying a constraint to refine these vectors. An examples of false memory and a correct one for the target class dog are shown in Fig.~\ref{fig:FM}. where our method can omit them. 

Moreover, we present several samples of retrieved images belong to several classes, when querying the learner network $\mathcal{L}$ using our novel memory recovery paradigm, as shown in Fig.~\ref{fig:impressions}. From the figure, we can see how the network retains its learned knowledge in its memory which is a specific pattern representing the target classes.
\pgfplotsset{every axis/.append style={
                    label style={font=\footnotesize},
                    tick label style={font=\scriptsize},
                    }}

    \begin{figure}[t]
        \begin{tikzpicture}
        \begin{groupplot}[group style={group size=2 by 1,
            horizontal sep=18pt,
            vertical sep=1cm},
            height=4cm,width=5cm,
        ]
        \nextgroupplot[
         xlabel=Transfer set size
        ,ylabel=Average Accuracy$(\%)$
        ,xtick=data
        ,ymax=100
        ,xmin=0
        ,xmax=4,
        ,ymin =0
        ,ymajorgrids
        ,xmajorgrids
        ,xticklabels={$\text{0}$,$\text{50}$,$\text{500}$,$\text{5000}$,$\text{6000}$},
        ]
        \addplot+[very thick,azure(colorwheel),mark size=1pt] coordinates
        {(0,19.241) (1,36.523) (2,59.416) (3,73.641) (4,75.326)} ;

        \nextgroupplot[
        legend style={font=\tiny}
         ,legend pos= south west
        ,xlabel=Number of classes
        ,xtick=data
        ,ymax=100
        ,xmin=0
        ,xmax=10,
        ,ymin =20
        ,ymajorgrids
        ,xmajorgrids
        ,xticklabels={$\text{0}$,$\text{2}$,$\text{4}$,$\text{6}$,$\text{8}$,$\text{10}$}
        ]
        \addplot+[thick,green(munsell),mark=Mercedes star,mark size=2pt] coordinates
        {(2,96.641) (4,78.432) (6,62.131) (8,52.152)(10,47.275)} ;

        \addplot+[thick,amaranth,mark=Mercedes star,mark size=2pt,mark size=2pt] coordinates
        {(2,96.735) (4,71.742) (6,58.263) (8,50.724)(10,42.834)} ;
        
        \legend{Real data, Retrieved data}
        
        \end{groupplot}
        \end{tikzpicture}
        \vspace{-2mm}
        \caption{Experimental results on CIFAR-10 dataset. (left) classification accuracy curves for ZS-IL with various transfer set size. (right) comparison between adopting original data versus recovered data in iCaRL\cite{rebuffi2017icarl} method.}
        \label{fig:vis2}
    \end{figure}
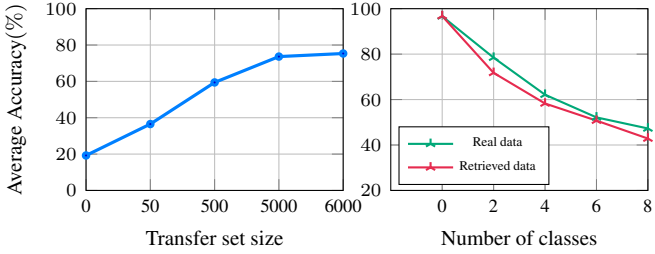
   
\textbf{Effect of transfer set size.} We examine the impact of transfer set size on the performance of the incrementally learning classes. To this end, we set up the proposed ZS-IL on CIFAR-10 dataset with different sizes of transfer sets, including $[50,500,5000,6000]$, wherein in the latest case, the retrieved data from the past is equal to the incremented class. Fig.~\ref{fig:vis2} (left) shows the performance. Obviously, increasing the number of synthesized samples in the transfer set has a significant impact on the performance. It is worth noting that an optimal size for the transfer set dependent on the task complexity in terms of the number of classes and variations of the actual images. Thus, increasing the set size higher than a reasonable one might increase the risk of overfitting.

\textbf{Zero-shot vs Few-Shot Incremental Learning.} To further evaluate the proposed method, we also investigate the Few-Shot Incremental Learning (FS-IL), where a few samples from the past are stored $\mathcal{M}$ to be integrated with the transfer set $\mathcal{S}$ when retraining the learner network $\mathcal{L}$ to learn a new incremented class $\mathcal{I}$. In this case, we store only $50$ samples for each class and our objective is to minimize the following cost function:
\begin{equation}
 \label{eq:4}
    \ell = \underset{x \in \mathcal{I}}{\ell_{CE}}(\mathcal{L}^{x}, y) + \lambda_{1} \underset{x \in \mathcal{S}}{\ell_{KD}}(\mathcal{L}^{x}, y) +  \lambda_{2}\underset{x \in \mathcal{M}}{\ell_{KD}}(\mathcal{L}^{x}, y) 
\end{equation}
where $\mathcal{L}^{x}$ stands for $\mathcal{L}(x;\theta_{\mathcal{L}})$, and $\lambda_1$ and $\lambda_2$  control the degree of distillation for each set. We report the result in Table~\ref{tab:FS}. Obliviously, the ablated model FS-IL outperforms the ZS-IL at the cost of requiring an external memory to save past examples. Obliviously, the ablated model FS-IL outperforms the ZS-IL at the cost of requiring an external memory to save past examples, which means our method is not only a standalone approach but also an extension to boost the performance of memory-based methods by providing a more balanced dataset at each time step.

\textbf{Hyper-parameters:$\lambda$ and $\eta$:}
 $\lambda$   control how much the transfer set $\mathcal{S}$ should be attended as Eq. \ref{eq:totalLoss}, and  $\eta$ is related to  how much the generated model output should be close to the recommendation vector. To select the best one for the two mentioned parameter, we perform grid search, where a brief of examination are reported in Fig.~\ref{fig:vis1}, where a take $\lambda$ and $\eta$ as $0.3$ and $0.45$, respectively.
 \begin{figure}[t]
        \begin{tikzpicture}
        \begin{groupplot}[group style={group size=2 by 1,
            horizontal sep=18pt,
            vertical sep=1cm},
            height=4cm,width=5cm,
        ]
        \nextgroupplot[
        legend style={font=\tiny,mark size=10pt},
        ,legend pos= south west
        ,xlabel=Number of classes
        ,ylabel=Average Accuracy$(\%)$
        ,xtick=data
        ,ymax=100
        ,xmin=0
        ,xmax=10,
        ,ymin =20
        ,ymajorgrids
        ,xmajorgrids
        ,xticklabels={$\text{0}$,$\text{2}$,$\text{4}$,$\text{6}$,$\text{8}$,$\text{10}$}
        ]
        \addplot[thick,amber] coordinates
        {(2,96.421) (4,79.153) (6,72.642) (8,63.642)(10,60.632)} ;
        
        \addplot[thick,amaranth] coordinates
        {(2,96.421) (4,82.252) (6,74.371) (8,69.175)(10,65.532)} ;
        
        \addplot[thick,airforceblue] coordinates
        {(2,96.231) (4,88.265) (6,80.754) (8,79.736)(10,75.346)} ;
        
        \addplot[thick,green(munsell)] coordinates
        {(2,96.121) (4,86.645) (6,78.221) (8,76.746)(10,73.436)} ;

        \legend{$\lambda=0.1$,$\lambda=0.2$,$\lambda=0.3$,$\lambda=0.4$}

        \nextgroupplot[
        legend style={ font=\tiny,mark size=10pt}
        ,legend pos= south west
        ,xlabel=Number of classes
        ,xtick=data
        ,ymax=100
        ,xmin=0
        ,xmax=10,
        ,ymin =20
        ,ymajorgrids
        ,xmajorgrids
        ,xticklabels={$\text{0}$,$\text{2}$,$\text{4}$,$\text{6}$,$\text{8}$,$\text{10}$}
        ]
        \addplot[thick,amber] coordinates
        {(2,96.421) (4,85.212) (6,79.542) (8,74.642)(10,70.023)} ;
        
        \addplot[thick,amaranth] coordinates
        {(2,96.326) (4,90.753) (6,82.698) (8,80.076)(10,74.641)} ;
        
        \addplot[thick,airforceblue] coordinates
       {(2,96.421) (4,80.252) (6,71.371) (8,69.175)(10,66.532)} ;
        
        \addplot[thick,green(munsell)] coordinates
       {(2,96.741) (4,78.352) (6,70.174) (8,67.365)(10,62.032)} ;

        \legend{$\eta=0.4$,$\eta=0.45$,$\eta=0.5$,$\eta=0.55$}
        
        \end{groupplot}
        \end{tikzpicture}
        \caption{Performance (average accuracy) comparison of the ZS-IL with various hyperparameters $\lambda$ (left) and $\eta$ (right).}
        \label{fig:vis1}
    \end{figure}
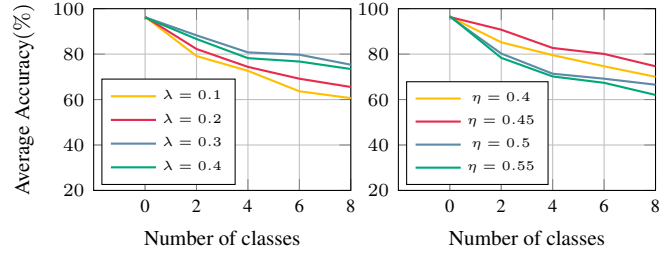
  \begin{table}[h]
\begin{center}
\caption{Classification results for CIFAR-10 in the Zero-Shot IL (ZS-IL) and Few-Shot IL (FS-IL) settings.}
\begin{tabular}{|c|cc|cc|}
\hline
\multirow{2}{*}{method} & \multicolumn{2}{c|}{CIFAR-10} & \multicolumn{2}{c|}{Tiny-ImageNet} \\ \cline{2-5} 
 & Class-IL & Task-IL & Class-IL & Task-IL \\ \hline
ZS-IL & 75.34 & 93.12 & 33.06 & 67.42 \\
FS-IL & 79.36 & 95.12 & 34.72 & 68.73 \\ \hline
\end{tabular}
\label{tab:FS} 
\end{center}
\end{table}
\vspace{-1mm}  

\textbf{ZS-IL in memory-based works}
Our suggested method is a better alternative to the buffer-based methods to omit the need for a memory buffer and decrease the risk of overfitting due to the more balanced fine-tuning at the same time. To validate this assertion, we embed our memory recovery paradigm into a prominent method iCaRL\cite{rebuffi2017icarl}. Performance result are shown in Fig.~\ref{fig:vis2} (right). From the figure, we can see adopting our ZS-IL can compromise between performance and memory footprint.
\vspace{-2mm}
\section{Conclusion}
\vspace{-2mm}
In this paper, we have proposed a novel strategy for incremental learning to address the memory issue, which is crucial when the number of classes becomes large. In particular, we perform incremental learning in both class-IL and task-IL settings in a zero-shot manner. This strategy is implemented through a memory recovery paradigm with no additional equipment. It only relies on the single DNN, known as the learner, to retrieve the network's past knowledge as a transfer set to look back on learned experiences. 
Our method has outstanding results on two challenging datasets CIFAR-10 and Tiny-ImageNet, compared with recent prominent works. To better show off the power of ZS-IL, we perform a clear and extensive comparison of SOTA methods considering both data-free and memory-based approaches.


{\small
\bibliographystyle{ieee_fullname}
\bibliography{Arxiv}
}

\end{document}